\def\year{2019}\relax
\documentclass[letterpaper]{article} 
\usepackage{aaai19}  
\usepackage{times}  
\usepackage{helvet}  
\usepackage{courier}  
\usepackage{url}  
\usepackage{graphicx}  
\usepackage[space]{cite}
\usepackage{xcolor}
\usepackage{float}
\usepackage{multirow}
\makeatletter
    \let\@internalcite\cite
    \def\cite{\def\citeauthoryear##1##2{##1, ##2}\@internalcite}
    \def\shortcite{\def\citeauthoryear##1{##2}\@internalcite}
    \def\@biblabel#1{\def\citeauthoryear##1##2{##1, ##2}[#1]\hfill}
\makeatother
\begin{document}
%
\title{Frustratingly Poor Performance of Reading Comprehension Models on Non-adversarial Examples}

\author{Soham Parikh, Ananya B. Sai, Preksha Nema, Mitesh M. Khapra\\
Indian Institute of Technology, Madras \\
\{sohamp,ananyasb,preksha,miteshk\}@cse.iitm.ac.in
}
\maketitle
\begin{abstract}
When humans learn to perform a difficult task (say, reading comprehension (RC) over \textit{longer} passages), it is typically the case that their performance improves significantly on an easier version of this task (say, RC over \textit{shorter} passages). Ideally, we would want an intelligent agent to also exhibit such a behavior. However, on experimenting with state of the art RC models using the standard RACE dataset, we observe that this is not true. Specifically, we see counter-intuitive results wherein even when we show frustratingly easy examples to the model at test time, there is hardly any improvement in its performance. We refer to this as \textit{non-adversarial} evaluation as opposed to adversarial evaluation. Such non-adversarial examples allow us to assess the utility of specialized neural components. For example, we show that even for easy examples where the answer is clearly embedded in the passage, the neural components designed for paying attention to relevant portions of the passage fail to serve their intended purpose. We believe that the non-adversarial dataset created as a part of this work would complement the research on adversarial evaluation and give a more realistic assessment of the ability of RC models.
All the datasets and codes developed as a part of this work will be made publicly available.
\end{abstract}

The availability of large-scale datasets for RC \cite{DBLP:journals/corr/HillBCW15, DBLP:conf/emnlp/OnishiWBGM16, DBLP:conf/emnlp/RajpurkarZLL16, DBLP:journals/corr/TrischlerWYHSBS16, DBLP:conf/nips/NguyenRSGTMD16} over the past few years has led to the development of increasingly complex neural models \cite{DBLP:journals/corr/XiongZS16,DBLP:conf/acl/CuiCWWLH17,DBLP:conf/acl/WangYWCZ17,DBLP:journals/corr/HuPQ17,DBLP:journals/corr/GongB17} with specialized components addressing various subtleties of this task. However, in parallel, research on adversarial evaluation \cite{DBLP:conf/emnlp/JiaL17, DBLP:journals/corr/LiangLSBLS17} suggests that these models essentially overfit on the training data and their performance drops drastically when presented with carefully constructed hard/confusing examples (referred to as \textit{adversarial} examples). In this work, we focus on the other end of this evaluation spectrum and study the performance of these models on carefully constructed easy examples at test time. We refer to this as \textit{non-adversarial evaluation}. 

The motivation for non-adversarial evaluation stems from the observation that when humans learn to perform a certain task of a certain difficulty level, they have the ability to naturally perform better on easier versions of the same task. For example, consider a student learning the task of RC who is able to perform at a certain level (say $k$\% accuracy). Now, if the same student is shown an easier version of this task (say, by providing hints pointing to the answer) then it is quite natural to expect his/her performance to improve by a considerable margin (\textit{i.e.}, he/she can now achieve an accuracy much greater than $k$\%).

If we can show that intelligent machines exhibit a similar behavior, we can say that they are closer to behaving like humans than otherwise. With this motivation, we perform a non-adversarial evaluation of state of the art RC models using the RACE dataset. This dataset contains triples taken from middle and high school English examinations of the form \textit{\{passage, query, options\}} and the task is to select the correct answer from the provided options. We propose different ways of constructing easy examples such as (i) embedding a declarative sentence formed by combining the query and the answer in the passage (ii) replacing the passage by the above mentioned declarative sentence (iii) appending the query with hints pointing to the correct option and so on. These examples are constructed in a way that an average high/middle school student would be able to answer them with almost 100\% accuracy. However, on experimenting with 5 state of the art RC models, we observe counter-intuitive results wherein there is hardly any improvement in the performance of these models on such easy examples (\textit{i.e.}, their performance remains at almost the same level as that on the standard RACE dataset). Of course, when such easy examples are added to the training data as well, then the performance of the model on the corresponding non-adversarial test set increases drastically ($95$-$99$\% accuracy) while the performance on the RACE dataset is almost the same as given by random guessing. This suggests that these models simply overfit the training data and do not learn any meaningful comprehension skills.

An added advantage of non-adversarial examples is that they allow us to easily investigate the role of some of the specialized modules of these complex neural RC models. For example, consider the module which is designed to pay attention to those portions of the passage which are relevant to the query and contain the answer. We observe that even when the query and answer are clearly embedded in the passage, this module fails to serve its intended purpose. We believe that carefully designing such non-adversarial evaluation strategies could lead to better explainability and accountability of RC models in particular and deep learning models in general. Specifically, given the general trend of adding increasingly complex modules to neural networks, it would help if, for any new module that is proposed, one could design such carefully constructed non-adversarial examples to check whether the module indeed serves its intended purpose (as opposed to just reporting numbers on a standard test set). In this context, the non-adversarial datasets created as a part of this work complements the research on adversarial evaluation and is well suited for getting a more realistic assessment of the abilities of RC models.

\section{Related Work}
This section is organized into $2$ parts (i) The first part introduces the datasets and models for RC QA (ii) The second part introduces adversarial evaluation. \\

\textbf{QA Datasets \& Models:}
Over the past few years, several large-scale datasets have been proposed for RC, inspiring increasingly complex models with many components. These datasets are of varied flavors and differ in whether the answer should be generated/selected/extracted. Cloze-style datasets like the CNN and Daily Mail \cite{DBLP:conf/nips/HermannKGEKSB15}, Children's Book Test \cite{DBLP:journals/corr/HillBCW15} and Who Did What \cite{DBLP:conf/emnlp/OnishiWBGM16} contain the answer inside the passage as an entity/verb/adjective, etc. SQuAD \cite{DBLP:conf/emnlp/RajpurkarZLL16}, TriviaQA \cite{DBLP:conf/acl/JoshiCWZ17} and NEWSQA \cite{DBLP:journals/corr/TrischlerWYHSBS16} require the RC model to predict a continuous span of text inside the passage as the answer. MSMARCO \cite{DBLP:conf/nips/NguyenRSGTMD16} contains human generated answers which in turn requires the models to perform answer generation. Other datasets like MCTest \cite{DBLP:conf/emnlp/RichardsonBR13}, NTCIR QA Lab Task \cite{DBLP:conf/ntcir/ShibukiSKMIIWMK14} and RACE \cite{DBLP:conf/emnlp/LaiXLYH17} contain multiple choice queries (MCQ) where the task is to select the correct option from a given set of options.

Most of the recent models proposed on these datasets have specialized attention modules like (i) Query-aware context representation \cite{DBLP:journals/corr/XiongZS16,DBLP:journals/corr/SeoKFH16,DBLP:conf/acl/ChenBM16,DBLP:conf/acl/CuiCWWLH17,DBLP:conf/acl/DhingraLYCS17} and 
(ii)Self-aware context representation \cite{DBLP:conf/acl/WangYWCZ17, DBLP:journals/corr/HuPQ17}. These attention-based modules aim at focusing on important passage words based on the information (\textit{i.e,} query or passage words) provided. In this work, we argue that it is important to assess the utility of these modules using non-adversarial examples.

\textbf{Adversarial Evaluation:} This line of work focuses on bringing out the poor generalization abilities of Deep Neural Networks by feeding them carefully constructed adversarial examples. \cite{DBLP:journals/corr/GoodfellowSS14} showed that by adding a small amount of carefully designed noise to an image, it is possible to fool image classification models into predicting a wrong label for the given image (even though there is no visible difference in the image after perturbation).

Adding noise to natural language text while preserving the semantics is tricky but some attempts have been made in this direction. \cite{DBLP:journals/corr/LiMJ16a} erase words from a discourse while trying to maintain the meaning of the text. \cite{zhao2018generating} use auto-encoders to map discrete text to continuous embeddings and then add perturbations in the continuous space. \cite{DBLP:conf/emnlp/JiaL17} use SQuAD dataset to show that if we insert adversarial sentences in the passage such that these sentences have a high word overlap with the answer sentence without affecting the answer, then the models get confused and the performance drops significantly. This exposes the vulnerability of the models in adversarial settings. In this paper, we explore the other end of the spectrum and show that when using non-adversarial examples which are significantly easier for humans to answer, there is hardly any improvement in the performance of existing RC models.

\section{Choice of Dataset}
As mentioned earlier, in this work, we focus on the RACE dataset. RACE is a large scale MCQ style dataset which gives us more scope for creating non-adversarial examples by suitably simplifying the (i) passage, (ii) query and/or (iii) options. More importantly, this is a hard dataset with a significant fraction of questions requiring reasoning and inference (refer to the \cite{DBLP:conf/emnlp/LaiXLYH17} for statistical details) where even state of the art models have been able to reach only 45\% accuracy while the human performance is about $90$\%. There is enough scope for the models to give an improved performance on non-adversarial examples over the actual test examples from RACE. In contrast, for some of the other large-scale datasets such as SQuAD and CNN/Daily Mail, state of the art models have already reached near human performance and hence there isn't enough scope for improvement in their performance on non-adversarial examples. There is a clear scope for designing such strategies for other flavors of QA such as (i) TriviaQA \cite{DBLP:conf/acl/JoshiCWZ17} which requires evidence from multiple passages and (ii) MS MARCO \cite{DBLP:conf/nips/NguyenRSGTMD16} which requires the generation of answers. Our work is an important first step in this direction and we hope that it will fuel interest in the development of such strategies for these other flavors of QA also. 

\section{Non-Adversarial Examples}
In this section, we describe different ways of creating non-adversarial examples. 
\subsection{Modifying the passage}
We propose different ways of modifying the passage to provide explicit or implicit hints about the answer.

\textbf{P1 - Append answer to the passage:} The simplest thing is to append the correct answer at the end of the passage. This is very naive and there is no reason why a model (or even a human for that matter) should be able to answer the query as the answer is placed out of context. However, we list it here for the sake of completeness.

\textbf{P2 - Append query \& answer to the passage:} If both the query and the answer are appended at the end of the passage then most humans would be easily able to answer the query without having to read this passage. This just requires very basic comprehension skill and we should expect a trained QA model to be able to find the answer. We can also think of the query as providing context for the answer.

\textbf{P3 - Append query, answer as a declarative sentence:} Building on the above intuition, to simplify things even further, we combine the query and the answer to form a sentence and append this sentence at the end of the passage. A majority of the queries in RACE are fill-in-the-blank style queries. It is straightforward to convert these query-answer pairs into a declarative sentence by simply replacing the blank with the answer. For other types of queries, we create declarative sentences  (refer to Figure \ref{fig:example_decl}) by using manually defined rules over CoreNLP constituency parses \footnote{We use the publicly available code provided by \cite{DBLP:conf/emnlp/JiaL17}}.

\textbf{P4 - Simplify the passage}
An alternative way to reduce the difficulty of the passages, we pass them through an automatic text simplification model \cite{DBLP:conf/acl/NisioiSPD17} and use the simplified passages during test time. 

\textbf{P5 - Retain only relevant sentences:} To remove extraneous information, we only retain those sentences of the passage which are needed to answer the query. In other words, the new passage is a query-specific summary of the original passage. Specifically, we collect these examples with the help of in-house annotators (proficient in English) who are shown the passage, query, and answer and are asked to retain only sentences from the passage which are required to answer the query.

\textbf{P6 - Replace passage by query \& answer:} This is an even more simplified version of \textbf{P2} wherein we replace the entire passage by the concatenation of query and answer. Typically, a simple bag-of-words model should also give a high performance on this dataset. 

\textbf{P7 - Replace passage by query \& answer as a declarative sentence:} Analogous to \textbf{P5}, this dataset is a simplified version of \textbf{P3} where the passage is replaced entirely by the declarative sentence formed by combining the query and the answer. Passages in \textbf{P5} and \textbf{P6} can again be considered to be variants of a query-specific summary of the passage and are embarrassingly simple and to the point. Most humans would be able to answer the queries with $100$\% accuracy.

\textbf{P8 - Replace passage with the answer:} The entire passage is replaced by the correct option. While the resulting passage hardly makes any syntactic and semantic sense, we include this for the sake of completeness. However, it is worth mentioning that if a human is given a triplet containing \{\textit{answer, query, options}\} instead of \{\textit{passage, query, options}\}, in the absence of any other information, the human will simply pick the answer (that is the most elementary thing to do), obtaining $100$\% accuracy. Again, a simple bag-of-words model would also be able to predict answers correctly with high accuracy.

\textbf{P9 - Place explicit hints to the answer:} We explicitly add the following text at the beginning of the passage ``The answer to \$QUERY\$ is at the end of the passage". Simultaneously, we append the following text at the end of the passage, ``The answer to \$QUERY\$ is \$ANSWER\$". \$QUERY\$ and \$ANSWER\$ are variables which are replaced by the actual query and the answer. Such hints make the task embarrassingly easy for humans, who would simply read the hint in the first sentence, skip reading the passage and pick the answer from the hint placed at the end.

\begin{figure}
\fbox{\begin{minipage}{23em}
\textbf{Passage:} Hidden in a small street in the south end of Springfield ... Frigo's is an Italian restaurant ... I stepped into Frigo's almost by accident when ... I have a feeling that I'll be picking up dinner for me and the kids at Frigo's soon. \textcolor{blue}{\textbf{(272 words)}}
\newline
\textbf{Query:}  How much did the writer pay for his first meal at Frigo's?
\newline
\textbf{Summarized Passage:} I stepped into Frigo's almost by accident when I had to stay in Springfield into the evening for an open house at the school where I work. I ordered the easiest meal possible: a chicken sandwich and a salad. It cost \$4.75 for the sandwich. The salad was \$4.99 and didn't have salad dressing on it. \textcolor{blue}{\textbf{(56 words)}}
\end{minipage}}
\caption{Example from \textbf{P4} dataset}
\label{fig:examplep4}
\end{figure}

\begin{figure}
\fbox{\begin{minipage}{23em}
\textbf{Query:}  What does the sentence ``You're quite a fellow to build this bridge!" mean?
\newline
\textbf{Answer:} John was great to build this bridge.
\newline
\textbf{Declarative Sentence:} The sentence ``You're quite a fellow to build this bridge!" means John was great to build this bridge.
\end{minipage}}
\caption{Example of declarative sentence created from query and answer}
\label{fig:example_decl}
\end{figure} 

\subsection{Modifying the query}
We take inspiration from school/college textbooks where tough queries are often appended with hints. On similar lines, we propose two simple modifications to the query.

\textbf{Q1 - Append answer as a hint to the query:} We append the text "Answer is \$ANSWER\$" at the end of the query. Such a hint just gives away the answer and it is extremely trivial for any human to answer such a query. 

\textbf{Q2 - Appending not-an-answer hint to the query :} Since the hint in \textbf{Q1} is too direct, in \textbf{Q2} we indicate the wrong options by appending the text "Answer is not \$OPTION1\$, \$OPTION2\$, \$OPTION3\$" to the query, assuming without loss of generality that the $4^{th}$ option is the correct option. Again, it would be very easy for a human to answer the query in the presence of such a hint.

\subsection{Modifying the options}
We propose the following ways of simplifying the options to make things easier for the model:

\textbf{O1 - Replace each option by query \& option as a declarative sentence:} We do not claim that this makes things dramatically easy for the model but the idea here is that the declarative sentence should help the model to read the query and options together in a better light.

\textbf{O2 - Replace wrong options with options from other example(s) :}
While the options are easier to read and comprehend in \textbf{O1}, the model has to still distinguish between confusing options. To simplify things, we replace the $3$ incorrect options in every example with randomly selected options from other examples. The idea is that since the $3$ wrong options are not relevant to the query (or the passage, in most cases), the reader should be able to assign low probability scores to these. Again, most humans would find this setup much easier and would be able to pick the right answer by elimination. We validate this by conducting human evaluations where the evaluators are only shown the query and $4$ options. They are able to reach a performance of $86.4\%$ on a test set of $500$ such examples without even reading the passage. 

\textbf{O3 - Reducing the number of options:} With four options, the chance of randomly guessing the answer is $25$\%. As we reduce the number of options the chance of randomly guessing improves ($33$\% with $3$ options and $50$\% with $2$ options). It would be interesting to see if there is a dramatic relative improvement in the performance of the model as compared to the random baseline when fewer options are provided. For example, with 4 options if the model gets $k$\% relative improvement w.r.t the random baseline of $25$\% then it would be interesting to see whether with reduced (2) options this relative improvement over the random baseline of $50$\% is greater than $k$\%. \newline
Moreover, instead of randomly dropping an incorrect option, we also create a test set where we ask in-house human annotators to select the most confusing incorrect option for a given tuple of \{\textit{passage, query, options}\}. This is done to check whether the performance of models is better in this case as opposed to randomly dropping an option. 

\section{Models employed} \label{Models_emp}
In the section, we describe the various state of the art models that we evaluated on the above non-adversarial examples. These models were originally proposed for SQuAD dataset wherein the task is to predict the correct span in the passage. They do not have components for encoding or selecting the options. Of these, \textit{viz.}, Gated Attention Reader (GAR) \cite{DBLP:conf/acl/DhingraLYCS17} and Stanford Attention Reader (SAR) \cite{DBLP:conf/acl/ChenBM16} have already been adapted for the RACE dataset by suitably modifying them to encode the options and select the right option instead of predicting a span). We refer the reader to the original RACE paper \cite{DBLP:conf/emnlp/LaiXLYH17} to see these modifications. In a similar vein, we suitably modify 3 other models and adapt them to the RACE dataset as described below.

\paragraph{Dynamic Co-attention Network (DCN)} This model \cite{DBLP:journals/corr/XiongZS16} consists of three modules: (i) document and query encoder, (ii) co-attention module and (iii) dynamic pointing decoder. First, we use a separate LSTM to encode the options and use the state of the LSTM at the last time-step as the vector representation of the option. The co-attention module used to pay attention over passage and query words simultaneously is used without any modifications for the RACE dataset. Lastly, we need to replace the dynamic pointing decoder which is used to predict the start and end locations of the answer span. We use the same modifications that \cite{DBLP:conf/emnlp/LaiXLYH17} proposed to adapt GAR and SAR for the RACE dataset. Specifically, we replace the output module by a simple bilinear attention layer which computes the bilinear similarity between document word representations and query representation (\textit{i.e.}, the final hidden state of the LSTM used for encoding the query).  We then normalize these {word, query} similarities using a softmax function to compute the attention weight for each passage word in light of the query. Using these weights, we use the weighted sum of passage words as the passage representation, which in turn is used to compute the bilinear similarity with the representation of each option. The option with the highest similarity score is predicted as the answer.

\paragraph{Bi-Directional Attention Flow} BiDAF is a very complex hierarchical multi-stage model containing 6 layers. We make some simplifications to this model for ease of experimentation and some modifications to adapt it to the RACE dataset. We do not use a character embedding layer and use a simple word embedding layer as opposed to the two-layer highway network in the original paper. For the contextual embedding layer described in the original paper, we use LSTMs for computing the representation of the passage and query and add another LSTM for computing the representations of the option. We retain the attention flow layer and the modeling layer as they are. To obtain a fixed length passage representation, a weighted\footnote{Attention weights computed as in Equation 3 in \cite{DBLP:journals/corr/SeoKFH16}} sum of passage word representations is computed. For predicting the answer, the method from the DCN description follows.

\paragraph{Mnemonic Reader (MNR)} as proposed in \cite{DBLP:journals/corr/HuPQ17} does iterative alignment in two steps: ($1$) interactive alignment between query and document to generate query-aware passage representation (\textit{i.e.}, a representation which focuses on important parts of the passage in light of the query) and ($2$) self-alignment of the query-aware passage representation with itself to make the representation self-aware (\textit{i.e.}, to fuse information across the passage to capture long range dependency between words). Here again, we mainly change the output layer to compute attention weights\footnote{refer Equation 15 in \cite{DBLP:journals/corr/HuPQ17}} for a weighted sum of passage words. For predicting the answer, the method from the BiDAF and DCN descriptions follow.  

\begin{table}
\begin{center}
\resizebox{\columnwidth}{!}{%
\begin{tabular}{|c|c c c c c|} 
 \hline
 & \multicolumn{5}{ c|}{Models} \\
 \hline
 Dataset & GAR & SAR & DCN & BiDAF & MNR \\ 
 \hline 
 RACE & \textbf{44.08} & 43.3 & 41.75 & 41.75 & 41.97\\ 
 \hline
 P1 & 44.77 & \textbf{44.93} & 41.97 & 42.4 & 42.3\\ 
 \hline
 P2 & 44.69 & \textbf{45.26} & 42.1 & 42.36& 42.2\\ 
 \hline
 P3 & 44.63 & \textbf{44.91} & 42.06 & 42.38& 42.22\\
 \hline
 P4 & 36.56 & \textbf{43.04} & 32.18 & 41.73 & 32.30 \\
 \hline 
 P6 & 46.84 & \textbf{55.65} & 46.39 & 46.92 & 47.22 \\
 \hline
 P7 & 47.22 & \textbf{55.55} & 46.66 & 46.05 & 47.83 \\
 \hline
 P8 & 49.53 & \textbf{65.99} & 51.66 & 50.2 & 53.91 \\
 \hline
 P9 & 44.45 & \textbf{45.12} & 41.97 & 42.56 & 42.14 \\
 \hline
 Q1 & 42.46 & 42.54 & 41.61 & 40.64 & \textbf{42.62} \\
 \hline
 Q2 & 34.21 & \textbf{37.19} & 36.68 & 34.35 & 27.79 \\
 \hline
 O1 & 39.04 & 39.87 & \textbf{39.93} & 39.14 & 39.28 \\
 \hline
 O2 & \textbf{43.6} & 43.17 & 41 & 41.02 & 37.86 \\
 \hline
 O3(3) & \textbf{52.03} & 51.64 & 50.06 & 50.43 & 50.81 \\
 \hline
 O3(2) & 65.89 & \textbf{66.21} & 65.71 & 65.14 & 64.98 \\
 \hline
\end{tabular}
}
\end{center}
\caption{Results of the 5 models when trained on original RACE dataset and tested on the non-adversarial versions of it. O3(3) and O3(2) correspond to having 3 and 2 options respectively.}
\label{restab}
\end{table}

\section{Results}
We train each of the $5$ models described above on the training set of the RACE dataset. We tune the hyperparameters of these models using the validation set of the RACE dataset to give the best performance. For benchmarking, we report the performance of these models on the test set on the RACE dataset. We then create $13$ non-adversarial test sets from the test set of the RACE dataset (P1 to P7, Q1-Q2 and O1 to O3). The hypothesis is that if the model has truly learned Natural Language Understanding (NLU) then its performance should be much better on these non-adversarial datasets than on the original RACE test set (just as we expect most humans to excel on these non-adversarial test examples). The results of our experiments are summarized in Table \ref{restab}. We make a few observations from these results:\\

\textbf{$\bullet$} On \textbf{P2} and \textbf{P3}, where the answer is present in the document along with the relevant context, we hardly see any improvement in the performance as compared to that on the original RACE dataset.
\\
\textbf{$\bullet$} On \textbf{P5} and \textbf{P6}, where only the answer along with its relevant context is present, the improvements of $4$ out of $5$ models are marginal. While improvement is significant with SAR model, the performance is still just above $50$\%.\\
\textbf{$\bullet$} SAR, which is the most simplistic model, gives the best performance on $10$ out of $14$ datasets. This, in turn, indicates that the more complex models have overfit on the original RACE dataset and are poor at generalizing on non-adversarial examples. \\
\textbf{$\bullet$} For the dataset \textbf{O3} with $3$ and $2$ options, the relative improvements of the best performing models over random guessing are $57$\% and $32$\% respectively while the relative improvement over random guessing on RACE dataset (with $4$ options) is $76$\%. This suggests that removing possibly confusing options does not seem to simplify things for the models by great extent.

\begin{table}
\begin{center}
\resizebox{\columnwidth}{!}{%
\begin{tabular}{|c|c c c c c|} 
 \hline
 & \multicolumn{5}{ c|}{Models} \\
 \hline
 Dataset & GAR & SAR & DCN & BiDAF & MNR \\ 
 \hline 
 RACE & \textbf{41.6} & 39 & 39.2 & 40 & 39.2\\ 
 \hline
 P5  & 41.88 & 41.08 & 41.28 & 39.68 &\textbf{ 43.09 }\\ 
 \hline
 
\end{tabular}
}
\end{center}
\caption{Results of the 5 models when trained on original RACE and tested on a subset of RACE test set ($500$ examples) and on the corresponding \textbf{P5} created using this test set}
\label{p4restab}
\end{table}

We also evaluate models on the human annotated datasets \textit{i.e.,} \textbf{P5}, where annotators select only the sentences required to answer the query and \textbf{O3(H)}, where the most confusing option indicated by annotators is dropped. Since this test set is created from a subset ($500$ examples) of the original RACE test set, we compare the performance of models on \textbf{P5} to their performance on the corresponding original subset in Table \ref{p4restab}. For \textbf{O3(H)}, in Table \ref{o3hrestab}, we compare the performance with a test set consisting of the same examples where an incorrect option is randomly dropped from each example.

\begin{table}
\begin{center}
\resizebox{\columnwidth}{!}{%
\begin{tabular}{|c|c c c c c|} 
 \hline
 & \multicolumn{5}{ c|}{Models} \\
 \hline
 Dataset & GAR & SAR & DCN & BiDAF & MNR \\ 
 \hline 
 O3(3) & 49.1 & \textbf{51.3} & 50.3 & 47.29 & 50.5\\ 
 \hline
 O3(H)  & 47.49 & 48.1 & 47.09 & 49.5 &\textbf{49.9}\\ 
 \hline
 
\end{tabular}
}
\end{center}
\caption{Results of the 5 models when trained on original RACE and tested on \textbf{O3(3)} and \textbf{O3(H)}. \textbf{O3(3)} corresponds to having 3 options, with one randomly dropped while \textbf{O3(H)} corresponds to dropping the most confusing options as judged by human annotators}
\label{o3hrestab}
\end{table}

\paragraph{}
Note that the RACE dataset also has a natural easy-hard split because it contains questions from mid school and high school exams (mid-school being presumably easier). So we did another experiment where we train the model on high school examples and evaluate it on both middle and high school examples at test time. Here again, from Table \ref{trainhightab} we observe that the performance of the model on the mid-school test is in fact lower than its performance on the high school test set.

\begin{table}[H]
\begin{center}
\resizebox{\columnwidth}{!}{%
\begin{tabular}{|c|c c c c c|} 
 \hline
 & \multicolumn{5}{ c|}{Models} \\
 \hline
 Dataset & GAR & SAR & DCN & BiDAF & MNR \\ 
 \hline 
 RACE-H & \textbf{41.57} & 41.51 & 41.02 & 40.17 & 41.14\\ 
 \hline
 RACE-M & 36.56 & 39.62 & \textbf{39.9} & 38.23 & 39.62\\ 
 \hline
\end{tabular}
}
\end{center}
\caption{Results of the 5 models when trained on RACE-H dataset and tested on the RACE-M test set.}
\label{trainhightab}
\end{table}

Lastly, we also wanted to check what happens to the performance of the model when trained on non-adversarial examples. We use non-adversarial versions of the training data to train separate models for each non-adversarial training sets. We then evaluate the trained models on the corresponding non-adversarial test set (\textit{i.e.}, evaluate the model trained using P1 type modifications on the test set containing P1 type modifications). As seen in Table \ref{trainnarestab}, the performance on non-adversarial test set now improves drastically (close to $98$\%) whereas the performance on the RACE test set is close to that given by random guessing, thereby showing that these models are only capable of learning patterns in the training data and do not exhibit any NLU. 
\begin{table}
\begin{center}
\resizebox{\columnwidth}{!}{%
\begin{tabular}{|c|c c c c c|} 
 \hline
 & \multicolumn{5}{ c|}{Models} \\
 \hline
 Dataset & GAR & SAR & DCN & BiDAF & MNR \\ 
 \hline 
 P9 & 98.66 & \textbf{98.89} & 98.01 & 98.46 & 95.07\\ 
 \hline
 RACE & 29.07 & 27.04 & 27.64 & 28.25 & \textbf{29.45}\\ 
 \hline
 
\end{tabular}
}
\end{center}
\caption{Results of the 5 models when trained on \textbf{P9} and tested on the original RACE test set}
\label{trainnarestab}
\end{table}

\section{Discussions and Analysis}
Attention modules are an important component of all the models described in Section \ref{Models_emp}.  Query-aware attention modules use query information to select important passage words. Self-matching modules use information from other passage words to select important passage words. These attention weights are then used to compute a passage representation which pays more attention to these words. Most previous works only do a qualitative analyses of the weights learned by these modules using a handful of examples. In this section, we show how specific non-adversarial examples can be used to quantitatively analyze the performance of these components.

\begin{table}
\begin{center}
\begin{tabular}{|c|c|c|c|c|} 
 \hline
 \multicolumn{2}{|c|}{Model/Data} &$<$ UAA&$<$ UAQ & MRR\\
 \hline
 \multirow{2}{*}{GAR} &P2&28.64&37.66&0.0089\\
 \cline{2-5}
 &P6&56.08&43.92&0.289\\
 \hline
 \multirow{2}{*}{SAR} &P2&27.97&27.97&0.007\\
 \cline{2-5}
 &P6&48.24&51.76&0.324\\
 \hline
 \multirow{2}{*}{DCN} &P2&65.12&55.53&0.09\\
 \cline{2-5}
 &P6&51.44&48.56&0.388\\
 \hline
 \multirow{2}{*}{BiDAF} &P2&27.22&44.02&0.005\\
 \cline{2-5}
 &P6&42.93&57.07&0.21\\
 \hline
 \multirow{2}{*}{MNR} &P2&32.45&31.52&0.0124\\
 \cline{2-5}
 &P6&50.08&49.92&0.32\\
 \hline
\end{tabular}
\end{center}
\caption{$<$ UAA (Uniform Attention to Answer) and $<$ UAQ (Uniform Attention to Query) denote the percentage of examples with less than uniform attention weight assigned to the corresponding N-gram. MRR is the Mean Reciprocal Rank of the N-gram based on the total attention weight.}
\label{att_stats}
\end{table}

\subsection{Output Layer Attention} \label{outatt}
As described in section \ref{Models_emp}, the output layer of all the models uses information from the query to compute the attention weights (or importance) of all the passage words. We consider the non-adversarial datasets \textbf{P2} and \textbf{P6} in which the answer is present verbatim inside the passage. While ideally, we would want all the attention weight to be distributed on these answer words, other parts of the passage may also contain information relevant for answering the query. We consider that in the absence of any information the models must simply learn a uniform attention over all the passage words (\textit{i.e.}, the weight assigned to each word should be $\frac{1}{len(P)}$ where $len(P)$ is the total number of words in the passage). Now, if the model has truly learned to pay attention to important words then we expect these attention weights to be distributed in such a way that the answer ($A$) words get more than uniform attention. If we denote the passage as a sequence of words $p_{0}, p_{1}, ... , p_{k + 1},...,p_{k+len(A)}$ and the attention weights assigned by this module to the words as $\alpha_{1}, \alpha_{2}, ..., \alpha_{k+1},...,\alpha_{k+len(A)}$, we expect $\sum_{i=1}^{len(A)}\alpha_{k+i} > \frac{len(A)}{len(P)}$ \textit{i.e.,} we expect the total attention mass on answer words to be more than the attention mass on remaining words. We observe that for over $25$\% of the examples in \textbf{P2} and for over $50$\% of the examples in \textbf{P6}, the total attention weights assigned to answer words is less than what would have been assigned using uniform attention over all words. In the case of \textbf{P6} the length of the passage is very small and hence, there is no distraction as the entire passage is simply replaced by the query and the answer. This suggests that these complex attention components do not really learn to pay attention to relevant words.

We also do another quantitative analysis wherein we consider all $n$-grams in the passage which have the same length as the answer. We compute a score for each $n$-grams as the sum of the attention weights on all the words in the $n$-gram. We then rank these $n$-grams based on this score and compute the MRR of the answer $n$-grams in this ranked list. As shown in Table \ref{att_stats}, we observe that the MRR is significantly low for each model for the dataset \textbf{P2}. The MRRs on \textbf{P6} are higher due to shorter passage length.  

\subsection{Query-Aware Document Attention}
GAR, DCN, BiDAF and MNR, each compute an affinity matrix $M \in R^{len(P) \times len(Q)}$, where $M_{ij}$ represents the similarity between the passage word at index $i$ and query word at index $j$. This module aims to highlight words in the passage which are important for each query word. Again, we use the datasets \textbf{P2} and \textbf{P6}, where the query is embedded verbatim inside the passage. Following arguments similar to the ones presented in Section \ref{outatt}, it is natural to expect high attention weights to the query n-gram present in the passage. We rank each N-gram of length $len(Q)$ based on the Frobenius norm of the sub-matrix $M_{i:i+len(Q),:}$ $\forall i \in [0,len(P) - len(Q)]$ and compute the MRR of the query N-gram. From Table \ref{qstats}, we observe that for \textbf{P2}, this MRR for each model is considerably low even though the N-gram has an exact match with the query. For \textbf{P6}, the MRR is high as expected since there is no text except for the query and the answer. Interestingly, the MRR for the answer N-gram computed similarly is higher than that of the query N-gram in $3$ out of $4$ models on the dataset \textbf{P2}. This further raises the question whether the query-aware attention modules fully serve their purpose. 

\begin{table} 
\begin{center}
\resizebox{\columnwidth}{!}{%
\begin{tabular}{|c|c|c|c|c|c|c|c|c|} 
 \hline
  \multirow{2}{*}{Metric}& \multicolumn{2}{ c|}{GAR} & \multicolumn{2}{ c|}{DCN} & \multicolumn{2}{ c|}{BiDAF} & \multicolumn{2}{ c|}{MNR} \\ 
 \cline{2-9}
  &P2&P6&P2&P6&P2&P6&P2&P6\\
 \hline 
 MRR(Q) &0.041& 0.561 &0.025 & 0.57 &0.024&0.503&0.08&0.72\\ 
 \hline
 MRR(A) & 0.06&0.363&0.09&0.492&0.058&0.497&0.048&0.25\\ 
 \hline
\end{tabular}
}
\end{center}
\caption{MRR for query (MRR(Q)) and answer (MRR(A)) N-grams in Query-Aware Passage Attention layer}
\label{qstats}
\end{table}

\subsection{Self-Matching Attention}
MNR computes a self-affinity matrix $M_{self} \in R^{len(P) \times len(P)}$, where $M_{ij}$ is the mutual affinity between words at indices $i$ and $j$ in the passage. While this is useful in multi-sentence reasoning, it is difficult to annotate the related sentences in all the passages for the original dataset. Instead, we use dataset \textbf{P9}, the first and the last sentences of each passage are related to a good extent. Recall that in \textbf{P9}, the first sentence contains the query and a hint that the answer is located in the last sentence. The last sentence in turn contains the query as well as the answer. We rank each N-gram of length $K$ in the passage (except for the N-grams in the first sentence) based on the Frobenius norm of the sub-matrix $M_{0:N-1,i:i+K-1}$ $\forall i \in [0,len(P) - K]$, where $N$ and $K$ are the lengths of the first and last sentences, and compute the MRR of the last sentence (as an N-gram of length $K$). The MRR we get is $0.496$, which indicates that $M_{self}$ indeed gives a high attention to sentences with overlapping words and serves its purpose. However, poor performance indicates there is still a lot to be desired in terms of NLU.

\section{Conclusion and Future Work}
We propose methods for generating non-adversarial examples for evaluating RC models and to the best of our knowledge, this is the first step in this direction. The failure to perform well and generalize on these examples supports the argument that existing RC models do not really exhibit any NLU but simply do pattern matching and overfit on the given data. Using specific non-adversarial examples, we propose methods to quantify the effectiveness of intermediate attention modules. We hope that our work will further encourage (i) creation of non-adversarial examples on other datasets (ii) methods for quantitatively analyzing other RC modules like multi-hop and multi-perspective matching and (iii) design of RC models with better natural language abilities.

\bibliographystyle{aaai}
\bibliography{aaai_bib}

\begin{thebibliography}{}

\bibitem[\protect\citeauthoryear{Chen, Bolton, and
  Manning}{2016}]{DBLP:conf/acl/ChenBM16}
Chen, D.; Bolton, J.; and Manning, C.~D.
\newblock 2016.
\newblock A thorough examination of the cnn/daily mail reading comprehension
  task.
\newblock In {\em {ACL} {(1)}}.
\newblock The Association for Computer Linguistics.

\bibitem[\protect\citeauthoryear{Cui \bgroup et al\mbox.\egroup
  }{2017}]{DBLP:conf/acl/CuiCWWLH17}
Cui, Y.; Chen, Z.; Wei, S.; Wang, S.; Liu, T.; and Hu, G.
\newblock 2017.
\newblock Attention-over-attention neural networks for reading comprehension.
\newblock In {\em {ACL} {(1)}},  593--602.
\newblock Association for Computational Linguistics.

\bibitem[\protect\citeauthoryear{Dhingra \bgroup et al\mbox.\egroup
  }{2017}]{DBLP:conf/acl/DhingraLYCS17}
Dhingra, B.; Liu, H.; Yang, Z.; Cohen, W.~W.; and Salakhutdinov, R.
\newblock 2017.
\newblock Gated-attention readers for text comprehension.
\newblock In {\em {ACL} {(1)}},  1832--1846.
\newblock Association for Computational Linguistics.

\bibitem[\protect\citeauthoryear{Gong and
  Bowman}{2017}]{DBLP:journals/corr/GongB17}
Gong, Y., and Bowman, S.~R.
\newblock 2017.
\newblock Ruminating reader: Reasoning with gated multi-hop attention.
\newblock {\em CoRR} abs/1704.07415.

\bibitem[\protect\citeauthoryear{Goodfellow, Shlens, and
  Szegedy}{2014}]{DBLP:journals/corr/GoodfellowSS14}
Goodfellow, I.~J.; Shlens, J.; and Szegedy, C.
\newblock 2014.
\newblock Explaining and harnessing adversarial examples.
\newblock {\em CoRR} abs/1412.6572.

\bibitem[\protect\citeauthoryear{Hermann \bgroup et al\mbox.\egroup
  }{2015}]{DBLP:conf/nips/HermannKGEKSB15}
Hermann, K.~M.; Kocisk{\'{y}}, T.; Grefenstette, E.; Espeholt, L.; Kay, W.;
  Suleyman, M.; and Blunsom, P.
\newblock 2015.
\newblock Teaching machines to read and comprehend.
\newblock In {\em {NIPS}},  1693--1701.

\bibitem[\protect\citeauthoryear{Hill \bgroup et al\mbox.\egroup
  }{2015}]{DBLP:journals/corr/HillBCW15}
Hill, F.; Bordes, A.; Chopra, S.; and Weston, J.
\newblock 2015.
\newblock The goldilocks principle: Reading children's books with explicit
  memory representations.
\newblock {\em CoRR} abs/1511.02301.

\bibitem[\protect\citeauthoryear{Hu, Peng, and
  Qiu}{2017}]{DBLP:journals/corr/HuPQ17}
Hu, M.; Peng, Y.; and Qiu, X.
\newblock 2017.
\newblock Mnemonic reader for machine comprehension.
\newblock {\em CoRR} abs/1705.02798.

\bibitem[\protect\citeauthoryear{Jia and Liang}{2017}]{DBLP:conf/emnlp/JiaL17}
Jia, R., and Liang, P.
\newblock 2017.
\newblock Adversarial examples for evaluating reading comprehension systems.
\newblock In {\em {EMNLP}},  2021--2031.
\newblock Association for Computational Linguistics.

\bibitem[\protect\citeauthoryear{Joshi \bgroup et al\mbox.\egroup
  }{2017}]{DBLP:conf/acl/JoshiCWZ17}
Joshi, M.; Choi, E.; Weld, D.~S.; and Zettlemoyer, L.
\newblock 2017.
\newblock Triviaqa: {A} large scale distantly supervised challenge dataset for
  reading comprehension.
\newblock In {\em {ACL} {(1)}},  1601--1611.
\newblock Association for Computational Linguistics.

\bibitem[\protect\citeauthoryear{Lai \bgroup et al\mbox.\egroup
  }{2017}]{DBLP:conf/emnlp/LaiXLYH17}
Lai, G.; Xie, Q.; Liu, H.; Yang, Y.; and Hovy, E.~H.
\newblock 2017.
\newblock {RACE:} large-scale reading comprehension dataset from examinations.
\newblock In Palmer, M.; Hwa, R.; and Riedel, S., eds., {\em Proceedings of the
  2017 Conference on Empirical Methods in Natural Language Processing, {EMNLP}
  2017, Copenhagen, Denmark, September 9-11, 2017},  785--794.
\newblock Association for Computational Linguistics.

\bibitem[\protect\citeauthoryear{Li, Monroe, and
  Jurafsky}{2016}]{DBLP:journals/corr/LiMJ16a}
Li, J.; Monroe, W.; and Jurafsky, D.
\newblock 2016.
\newblock Understanding neural networks through representation erasure.
\newblock {\em CoRR} abs/1612.08220.

\bibitem[\protect\citeauthoryear{Liang \bgroup et al\mbox.\egroup
  }{2017}]{DBLP:journals/corr/LiangLSBLS17}
Liang, B.; Li, H.; Su, M.; Bian, P.; Li, X.; and Shi, W.
\newblock 2017.
\newblock Deep text classification can be fooled.
\newblock {\em CoRR} abs/1704.08006.

\bibitem[\protect\citeauthoryear{Nguyen \bgroup et al\mbox.\egroup
  }{2016}]{DBLP:conf/nips/NguyenRSGTMD16}
Nguyen, T.; Rosenberg, M.; Song, X.; Gao, J.; Tiwary, S.; Majumder, R.; and
  Deng, L.
\newblock 2016.
\newblock {MS} {MARCO:} {A} human generated machine reading comprehension
  dataset.
\newblock In {\em CoCo@NIPS}, volume 1773 of {\em {CEUR} Workshop Proceedings}.
\newblock CEUR-WS.org.

\bibitem[\protect\citeauthoryear{Nisioi \bgroup et al\mbox.\egroup
  }{2017}]{DBLP:conf/acl/NisioiSPD17}
Nisioi, S.; Stajner, S.; Ponzetto, S.~P.; and Dinu, L.~P.
\newblock 2017.
\newblock Exploring neural text simplification models.
\newblock In {\em Proceedings of the 55th Annual Meeting of the Association for
  Computational Linguistics, {ACL} 2017, Vancouver, Canada, July 30 - August 4,
  Volume 2: Short Papers},  85--91.

\bibitem[\protect\citeauthoryear{Onishi \bgroup et al\mbox.\egroup
  }{2016}]{DBLP:conf/emnlp/OnishiWBGM16}
Onishi, T.; Wang, H.; Bansal, M.; Gimpel, K.; and McAllester, D.~A.
\newblock 2016.
\newblock Who did what: {A} large-scale person-centered cloze dataset.
\newblock In {\em {EMNLP}},  2230--2235.
\newblock The Association for Computational Linguistics.

\bibitem[\protect\citeauthoryear{Rajpurkar \bgroup et al\mbox.\egroup
  }{2016}]{DBLP:conf/emnlp/RajpurkarZLL16}
Rajpurkar, P.; Zhang, J.; Lopyrev, K.; and Liang, P.
\newblock 2016.
\newblock Squad: 100, 000+ questions for machine comprehension of text.
\newblock In {\em {EMNLP}},  2383--2392.
\newblock The Association for Computational Linguistics.

\bibitem[\protect\citeauthoryear{Richardson, Burges, and
  Renshaw}{2013}]{DBLP:conf/emnlp/RichardsonBR13}
Richardson, M.; Burges, C. J.~C.; and Renshaw, E.
\newblock 2013.
\newblock Mctest: {A} challenge dataset for the open-domain machine
  comprehension of text.
\newblock In {\em {EMNLP}},  193--203.
\newblock {ACL}.

\bibitem[\protect\citeauthoryear{Seo \bgroup et al\mbox.\egroup
  }{2016}]{DBLP:journals/corr/SeoKFH16}
Seo, M.~J.; Kembhavi, A.; Farhadi, A.; and Hajishirzi, H.
\newblock 2016.
\newblock Bidirectional attention flow for machine comprehension.
\newblock {\em CoRR} abs/1611.01603.

\bibitem[\protect\citeauthoryear{Shibuki \bgroup et al\mbox.\egroup
  }{2014}]{DBLP:conf/ntcir/ShibukiSKMIIWMK14}
Shibuki, H.; Sakamoto, K.; Kano, Y.; Mitamura, T.; Ishioroshi, M.; Itakura,
  K.~Y.; Wang, D.; Mori, T.; and Kando, N.
\newblock 2014.
\newblock Overview of the {NTCIR-11} qa-lab task.
\newblock In {\em {NTCIR}}.
\newblock National Institute of Informatics {(NII)}.

\bibitem[\protect\citeauthoryear{Trischler \bgroup et al\mbox.\egroup
  }{2016}]{DBLP:journals/corr/TrischlerWYHSBS16}
Trischler, A.; Wang, T.; Yuan, X.; Harris, J.; Sordoni, A.; Bachman, P.; and
  Suleman, K.
\newblock 2016.
\newblock Newsqa: {A} machine comprehension dataset.
\newblock {\em CoRR} abs/1611.09830.

\bibitem[\protect\citeauthoryear{Wang \bgroup et al\mbox.\egroup
  }{2017}]{DBLP:conf/acl/WangYWCZ17}
Wang, W.; Yang, N.; Wei, F.; Chang, B.; and Zhou, M.
\newblock 2017.
\newblock Gated self-matching networks for reading comprehension and question
  answering.
\newblock In {\em {ACL} {(1)}},  189--198.
\newblock Association for Computational Linguistics.

\bibitem[\protect\citeauthoryear{Xiong, Zhong, and
  Socher}{2016}]{DBLP:journals/corr/XiongZS16}
Xiong, C.; Zhong, V.; and Socher, R.
\newblock 2016.
\newblock Dynamic coattention networks for question answering.
\newblock {\em CoRR} abs/1611.01604.

\bibitem[\protect\citeauthoryear{Zhengli~Zhao}{2018}]{zhao2018generating}
Zhengli~Zhao, Dheeru~Dua, S.~S.
\newblock 2018.
\newblock Generating natural adversarial examples.
\newblock {\em International Conference on Learning Representations}.

\end{thebibliography}
\end{document}